\documentclass[letterpaper]{article}
\usepackage{arxiv}
\usepackage[hyphens]{url}
\usepackage{graphicx}
\usepackage{natbib}
\usepackage{caption}
\usepackage{booktabs}
\usepackage[table]{xcolor}
\usepackage{multirow}
\usepackage{amsmath}
\usepackage{amssymb}
\usepackage{algorithm}
\usepackage{algorithmic}

\title{ResKV: Reconstructing Omitted Attention Contributions for Fixed-Budget
KV Cache Compression}
\author{
Yuhang Zhan,
Lisi Chen,
Shuo Shang
}
\affiliations{
University of Electronic Science and Technology of China\\
agooostoo@gmail.com, lchen012@e.ntu.edu.sg, jedi.shang@gmail.com
}

\begin{document}

\maketitle

\begin{abstract}
KV cache compression is essential for efficient long-context inference.
Existing eviction methods permanently discard unselected tokens and
consequently remove their aggregate contribution to attention. Merging-based
alternatives preserve more information but can perturb retained keys and
values that should remain exact. We observe that the information omitted by
cache eviction can be formulated as residual statistics in both the numerator
and denominator of softmax attention. Based on this observation, we propose
\textbf{ResKV}, which divides a fixed KV budget into an exact main cache and a
compact residual cache that reconstructs the contribution of omitted tokens.
ResKV lets main-cache tokens and residual entries participate in the same
softmax normalization, so residual entries restore both attention numerator and
denominator mass rather than acting as a post-hoc correction. A construction-time
validation proxy determines residual allocation for each layer and KV head,
while a decode-time dynamic gate adjusts residual contributions for individual
queries.
Comprehensive evaluations on LongBench and RULER, covering query-aware and
query-agnostic settings, multiple backbones, cache budgets, and representative
compression baselines, demonstrate broad improvements under the same
retained KV budget while preserving the practical efficiency of compressed
decoding, including peak memory usage and long-context decode throughput.
\end{abstract}

\section{Introduction}

Long-context large language models rely on a KV cache to avoid recomputing
past keys and values during autoregressive generation. While this cache makes
decode efficient, it also grows linearly with the context length and quickly
becomes a dominant source of memory use and memory traffic~\citep{kwon2023pagedattention,sheng2023flexgen,liu2024kivi,hooper2024kvquant}. This creates a
central tension in long-context inference: the model should retain enough
historical information to answer future queries, but the system can only afford
to keep a limited number of KV slots.

KV cache compression methods commonly address this tension by deciding which
tokens to keep. Eviction-based methods score cached tokens, retain a subset as
exact keys and values, and permanently discard the rest~\citep{liu2023scissorhands,zhang2023h2o,adnan2024keyformer,li2024snapkv,oren2024tova,tang2025razorattention,feng2025adakv,zhou2025dynamickv,gu2026obcache}. This design is
efficient, but it makes cache compression a binary decision: each token is either
kept unchanged or removed together with its value contribution and softmax
normalization mass.

Merging-based methods preserve information from evicted tokens
by combining related cache states. CaM folds evicted states into retained
entries, KVMerger merges similar keys with Gaussian-kernel weights, KeepKV
compensates attention scores using merging history, and SemantiCache constructs
semantic cores with proportional attention~\citep{zhang2024cam,wang2024kvmerger,tian2026keepkv,wu2026semanticache}.
These methods show that omitted information can remain useful after aggregation.
However, they fold omitted content into retained or merged cache states, which can
perturb retained memories and erase the individual identities of folded tokens.

Another line of recent work estimates the contribution of compressed or omitted
tokens. KVSculpt optimizes unconstrained virtual KV
pairs to preserve layer attention behavior~\citep{jiang2026kvsculpt}, while RESA
uses a low-rank logit prior and an online aggregator to compensate the output of
sparse attention~\citep{yang2026resa}. However, these methods do not jointly
address how to preserve selected
tokens as exact memories and represent omitted tokens as cache-resident entries
under the same fixed budget.

We instead view the removed tokens through the attention computation itself. In the
attention numerator--denominator form, the evicted side is a residual
contribution: a pair of omitted softmax statistics over the tokens outside the
main cache. If these omitted statistics could be reconstructed, the model would
not need to choose between deleting them outright, folding them into retained
tokens, or replacing them with merged cache states. The key question is
therefore how to represent this residual contribution within the same fixed KV
budget.

We propose \textbf{ResKV}, a residual KV cache that splits a fixed cache budget
into two parts: a main cache that stores selected tokens exactly, and a
residual cache that represents the omitted side with compact residual entries.
The main cache keeps high-priority tokens unchanged as exact memories, while
the residual cache reintroduces structured information from evicted tokens
without increasing the total number of cache slots. In this way, ResKV changes
the cache representation from a pure keep-or-drop decision into a
main-plus-residual layout under the same retained KV budget. This avoids the
binary deletion of eviction, does not fold omitted information into exact
retained memories, and keeps the omitted side as entries that can participate in
attention.

At decode time, ResKV combines the exact main cache and the residual cache
through a shared-softmax residual formulation. Residual entries are not used
as a post-hoc output correction; instead, they participate in the same softmax
normalization as the exact main-cache tokens. This lets each residual entry
restore both numerator mass and denominator mass, making it an approximate
attention participant rather than an external value update. ResKV further uses
adaptive residual control at two points. During cache construction, a
validation proxy selects the residual budget for each layer and KV head only
when residual entries improve held-out attention-output reconstruction. During decode, a
dynamic gate scales the residual logits according to the sharpness of the
main-cache attention, allowing residual mass under diffuse attention while
protecting sharp retrieval peaks.

We evaluate ResKV across multiple long-context benchmarks, backbones, and
representative compression baselines.
We consider both query-aware and query-agnostic construction; the latter
compresses the cache without access to the future query and therefore better
matches realistic deployment.
Across the main tables, ResKV improves all 32 displayed LongBench
configurations and 63 of the 64 displayed RULER configurations under the same
retained KV budget, with especially clear benefits under tight cache budgets
and on tasks that require distributed contextual evidence, with negligible
additional memory overhead and stable throughput at long
context lengths. Ablations further confirm the roles of the validation proxy,
dynamic gate, and shared softmax.

Our contributions are threefold:
\begin{itemize}
    \item We introduce ResKV, a main-plus-residual KV cache representation that
    accounts for omitted attention mass under a fixed KV budget while keeping
    selected main-cache entries exact.
    \item We develop adaptive residual control for ResKV, including a
    construction-time validation proxy and a decode-time dynamic gate.
    \item We show through comprehensive experiments that ResKV improves
    representative compression baselines across long-context benchmarks,
    cache budgets, and construction settings, while preserving practical efficiency.
\end{itemize}

\section{Related Work}

\paragraph{KV cache eviction and budget allocation.}
KV cache eviction exploits attention sparsity by retaining a subset of past
tokens. Existing methods preserve attention sinks and recent tokens, select
entries from accumulated or observed attention, adapt budgets across layers or
heads, or use removal-induced perturbation as saliency~\citep{xiao2024streamingllm,liu2023scissorhands,zhang2023h2o,adnan2024keyformer,li2024snapkv,oren2024tova,cai2025pyramidkv,tang2025razorattention,wang2025squeezeattention,feng2025adakv,zhou2025dynamickv,gu2026obcache}.

\paragraph{Representing omitted cache information.}
Several methods go beyond permanent deletion by encoding omitted content into
compact representations. CaM, KVMerger, KeepKV, and SemantiCache merge or
aggregate evicted states using attention importance, key similarity, merging
history, or semantic cores~\citep{zhang2024cam,wang2024kvmerger,tian2026keepkv,wu2026semanticache}.
KVSculpt distills compressed regions into unconstrained virtual KV pairs to
preserve attention behavior~\citep{jiang2026kvsculpt}. RESA estimates omitted
contributions with a low-rank logit prior and compensates sparse-attention
outputs online~\citep{yang2026resa}. LESS augments sparse attention with a learned recurrent residual state, and
ClusterKV organizes cache access around semantic clusters~\citep{dong2024less,liu2025clusterkv}.

\paragraph{Orthogonal KV cache optimizations.}
KV cache quantization reduces per-entry precision, while sparse-attention and
retrieval systems reduce the entries accessed at each step without necessarily
discarding the full cache~\citep{hooper2024kvquant,liu2024kivi,kang2024gear,sharma2025minikv,tang2024quest,liu2024retrievalattention}.

\section{Preliminaries}

\subsection{KV Cache}

An autoregressive Transformer~\citep{vaswani2017attention} processes a prompt in a \emph{prefill} phase and
then generates tokens one by one in a \emph{decode} phase. Prefill computes and
stores the prompt keys and values in a \emph{KV cache}; each decode step appends
the new token's key and value and attends with a single query $q$ over the
cached matrices,
\begin{equation}
o=\mathrm{softmax}\!\big(qK^\top/\sqrt{d}\big)\,V ,
\label{eq:attn}
\end{equation}
where $K,V\in\mathbb{R}^{s\times d}$ are the cached keys and values. The cache grows with
the sequence length and, at long context, dominates the memory and bandwidth of
inference~\citep{kwon2023pagedattention,sheng2023flexgen,liu2024kivi,hooper2024kvquant}.

\subsection{KV Cache Compression}

Under compression ratio $\rho\in(0,1)$, only
$b=\lfloor s(1-\rho)\rfloor$ KV slots are retained. \emph{Eviction} methods score
cached tokens, keep a set $M_0$ of size $b$, and drop
$E_0=S\setminus M_0$, so decoding attends only over the pruned cache,
\begin{equation}
o^{\mathrm{evict}}=\mathrm{softmax}\!\big(qK_{M_0}^\top/\sqrt{d}\big)\,V_{M_0},
\label{eq:evict}
\end{equation}
with $K_{M_0},V_{M_0}$ the retained rows~\citep{liu2023scissorhands,zhang2023h2o,adnan2024keyformer,li2024snapkv,oren2024tova,tang2025razorattention,gu2026obcache}.
\emph{Merging} methods instead fold evicted states back into retained slots, so
part of the omitted information remains within the same budget~\citep{zhang2024cam,wang2024kvmerger,tian2026keepkv}.

\section{Motivation}

\subsection{Observation}

\paragraph{Observation 1: hard eviction discards residual information.}
Hard eviction drops all terms from the evicted set, which is risky under diffuse
attention because many low-scoring tokens can still carry substantial aggregate mass
and may become useful again for future decoding queries.
To make this explicit, let the rows of $K$ and $V$ be $k_p$ and $v_p$, and define
$a_p=\langle q,k_p\rangle/\sqrt{d}$. Then
\begin{equation}
\begin{aligned}
o
&=\mathrm{softmax}\!\big(qK^\top/\sqrt{d}\big)V
 =\mathrm{softmax}\!\big([a_1,\ldots,a_s]\big)V \\
&=\sum_{p=1}^{s}\frac{e^{a_p}}{\sum_{p'=1}^{s}e^{a_{p'}}}\,v_p
 =\frac{\sum_{p=1}^{s} e^{a_p}v_p}{\sum_{p'=1}^{s} e^{a_{p'}}} \\
&=\frac{\sum_{p\in M_0}e^{a_p}v_p+\sum_{p\in E_0}e^{a_p}v_p}
        {\sum_{p'\in M_0}e^{a_{p'}}+\sum_{p'\in E_0}e^{a_{p'}}} .
\end{aligned}
\label{eq:expand}
\end{equation}
In this form, eviction deletes the two $E_0$ sums from the numerator and denominator,
then renormalizes using only the retained denominator over $M_0$. Thus, even if each
evicted token has small individual attention, their aggregate numerator and
normalization mass can still be missing from the decoding output.

\paragraph{Observation 2: folding the residual into exact tokens can corrupt them.}
Equation~\ref{eq:expand} also shows that the missing information is a separate pair
of sums over $E_0$. Merging avoids outright deletion, but it writes these evicted
terms into retained slots, changing the retained keys and values themselves. This is
undesirable for sharp retrieval: a token in $M_0$ that should remain an exact memory
can be perturbed by unrelated evicted tokens. These two failure modes suggest that the
missing information should be accounted for separately, without modifying the retained
cache.


\subsection{The Residual}

Equation~\ref{eq:expand} shows that the information lost by eviction is exactly a pair
of unnormalized softmax statistics. For a token subset $\mathcal{T}$ and query $q$,
with $a_p(q)=\langle q,k_p\rangle/\sqrt d$, define
\begin{equation}
Z_{\mathcal{T}}(q)=\sum_{p\in\mathcal{T}} e^{a_p(q)},\qquad
N_{\mathcal{T}}(q)=\sum_{p\in\mathcal{T}} e^{a_p(q)}\,v_p ,
\label{eq:zn}
\end{equation}
so that the full-cache output is $o_S(q)=N_S(q)/Z_S(q)$ and hard eviction gives the
main-only output $o_{M_0}(q)=N_{M_0}(q)/Z_{M_0}(q)$ (Eqs.~\ref{eq:attn}
and~\ref{eq:evict}).

Now consider a main cache that keeps a set $M$ exactly, with evicted complement
$E=S\setminus M$. We call the part omitted by the main cache the
\emph{residual} of $M$:
\begin{equation}
\begin{aligned}
\mathcal{R}(M;q)
&=\big(N_S(q)-N_M(q),\,Z_S(q)-Z_M(q)\big) \\
&=\big(N_E(q),\,Z_E(q)\big) \\
&=\left(\sum_{p\in E}e^{a_p(q)}v_p,\ \sum_{p\in E}e^{a_p(q)}\right).
\end{aligned}
\label{eq:residual}
\end{equation}
Adding $\mathcal{R}(M;q)$ back would recover full attention exactly, but it is not cacheable: the
weights $e^{a_p(q)}$ depend on the future query $q$, so evaluating it needs the evicted keys
and values. The core idea of ResKV is to trade exactness for reconstruction---keep the
few most important tokens exactly and use a compact residual cache to approximate
$\mathcal{R}(M;q)$ through attention-output reconstruction. Concretely, the budget is split as
$b=m+r$: a main cache $M$ of $m$ exact KV slots and a residual cache $R$ of $r$ slots,
leaving the retained tokens untouched. Conceptually, for fixed caches $M$ and $R$,
the best residual reconstruction $\widehat R$ would make residual-augmented attention
close to full attention:
\begin{equation}
\widehat R^\star
=\arg\min_{\widehat R}
\mathbb{E}_{q\sim\mathcal Q}\left\|
o_S(q)-\hat o(q)
\right\|_2^2 ,
\label{eq:residual_objective}
\end{equation}
where
\begin{equation}
o_S(q)=\frac{N_S(q)}{Z_S(q)},\qquad
\hat o(q)=
\frac{N_M(q)+\widehat N_{\widehat R}(q)}
     {Z_M(q)+\widehat Z_{\widehat R}(q)} .
\end{equation}
This form requires the residual contribution to share the same normalization as
the main cache, rather than being added as a separate output correction.
The next section describes how ResKV instantiates this attention-output
reconstruction within a fixed KV budget.

\section{Residual KV Cache (ResKV)}

\subsection{Overview}

\begin{figure*}[t]
\centering
\includegraphics[width=0.95\textwidth]{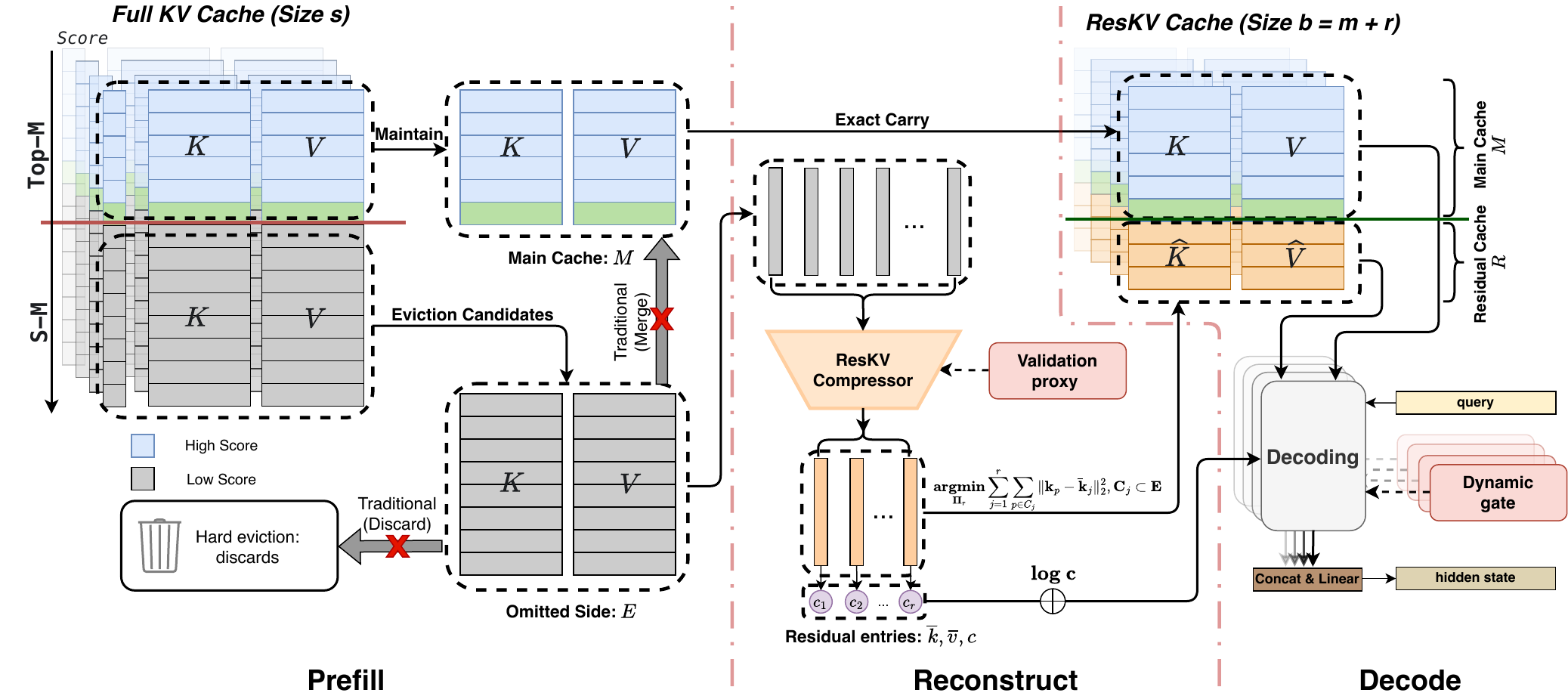}
\caption{Overview of ResKV. Given a fixed budget $b=m+r$, ResKV keeps selected
tokens exactly in the main cache $M$ and reconstructs the omitted side $E$ as a
residual cache $R$. During decoding, the main and residual branches are combined
by a dynamic gate.}
\label{fig:pipeline}
\end{figure*}

ResKV turns the residual view from the previous section into a main-plus-residual
cache layout. For each layer and KV head, the fixed budget is
$b=\lfloor s(1-\rho)\rfloor=m+r$: $m$ exact main-cache slots and $r$ residual
slots, keeping the same persistent KV cache footprint. Figure~\ref{fig:pipeline}
summarizes this layout and its decode-time combination.

Main entries remain exact, while residual entries summarize omitted tokens and
re-enter the shared softmax through their representative keys, values, and
counts. A validation proxy selects the residual allocation after prefill, and a
dynamic gate modulates residual logits during decode.

\subsection{Cache Split and Selection}

ResKV first splits the fixed cache budget $b$ into two parts, $b=m+r$. The main cache $M$
uses $m$ slots to store selected tokens with their original keys and values, so these
entries remain exact during decoding. The residual cache $R$ uses the remaining $r$ slots
for attention-output reconstruction of the omitted contribution. Given a per-token
importance score $s_p$ and a protected recent window $\mathcal{P}$, the main cache is
selected as
\begin{equation}
M=\mathcal{P}\cup\mathrm{Top}_{\,m-|\mathcal{P}|}(s),
\qquad
E=S\setminus M .
\label{eq:main_selection}
\end{equation}
where the recent window protects local continuity in autoregressive decoding. The
remaining prompt tokens form the evicted side $E$, whose contribution will be
represented by the residual cache and reintroduced during decoding.

\subsection{Residual Cache Construction}

Given the split $(M,E)$ and a residual budget $r$, ResKV approximates
$\mathcal R(M;q)$ by reconstructing the omitted statistics of $E$ for a future
query $q$, with
$a_p(q)=\langle q,k_p\rangle/\sqrt d$:
\begin{equation}
Z_E(q)=\sum_{p\in E} e^{a_p(q)},\qquad
N_E(q)=\sum_{p\in E} e^{a_p(q)}v_p .
\label{eq:omitted_stats}
\end{equation}
Because the weights depend on the future query, ResKV caches compact
query-agnostic summaries instead of the exact statistics.

Let $\Pi_r=\{C_1,\ldots,C_r\}$ be a partition of $E$. Each residual entry represents
one group $C_j$ and stores three quantities:
\begin{equation}
\bar{k}_j=\frac{1}{c_j}\sum_{p\in C_j} k_p,\quad
\bar{v}_j=\frac{1}{c_j}\sum_{p\in C_j} v_p,\quad
c_j=|C_j|,
\label{eq:summary}
\end{equation}
namely a representative key, a representative value, and a population count. If
the tokens in $C_j$ have similar logits for a query, $a_p(q)\approx \bar a_j(q)$ with
$\bar a_j(q)=\langle q,\bar k_j\rangle/\sqrt d$, then the group's omitted statistics
can be reconstructed as
\begin{equation}
\sum_{p\in C_j} e^{a_p(q)} \approx c_j e^{\bar a_j(q)},\qquad
\sum_{p\in C_j} e^{a_p(q)}v_p \approx c_j e^{\bar a_j(q)}\bar v_j .
\label{eq:entry_reconstruction}
\end{equation}
Thus the representative key and value reconstruct the group's address and
numerator, while $c_j$ restores its softmax mass. When $c_j=1$, a residual entry
reduces to an exact retained token.

Since future queries are unavailable at cache-construction time, ResKV uses
key-space consistency as a query-agnostic surrogate: nearby keys tend to induce
similar logits and can share one residual entry. We choose the partition by
minimizing
\begin{equation}
\Pi_r^\star
=\arg\min_{\Pi_r}
\sum_{j=1}^{r}\sum_{p\in C_j}\left\lVert k_p-\bar k_j\right\rVert_2^2,
\qquad C_j\subset E .
\label{eq:residual_partition}
\end{equation}
We solve this objective with a few Lloyd iterations~\citep{lloyd1982least},
initialized from high-scored tokens in $E$, independently for each layer and KV
head after prefill. The resulting $r$ residual entries occupy $r$ cache slots,
so the total budget remains $b=m+r$.

\subsection{Shared-Softmax Residual Decode}

At decode, exact main-cache tokens and residual entries participate in one
shared softmax rather than using the residual branch as a post-hoc correction.
For a query $q$, define the main logits
\begin{equation}
a_p=\langle q,k_p\rangle/\sqrt d,\qquad p\in M,
\end{equation}
and the residual-entry logits
\begin{equation}
\tilde a_j
=\langle q,\bar k_j\rangle/\sqrt d+\log c_j+\log g(q),
\qquad j\in R .
\label{eq:residual_logit}
\end{equation}
The term $\log c_j$ injects the population mass from
Eq.~\ref{eq:entry_reconstruction}, and $g(q)\in[0,1]$ is the dynamic gate
defined below.

We write the shared softmax in the same numerator--denominator form as the motivation:
\begin{align}
\widehat Z(q)
&=\sum_{p\in M}e^{a_p}+\sum_{j\in R}e^{\tilde a_j}, \label{eq:reskv_den}\\
\widehat N(q)
&=\sum_{p\in M}e^{a_p}v_p+\sum_{j\in R}e^{\tilde a_j}\bar v_j .
\label{eq:reskv_num}
\end{align}
The output is then
\begin{equation}
\hat o(q)=\frac{\widehat N(q)}{\widehat Z(q)} ,
\label{eq:reskv}
\end{equation}
where residual terms contribute to both the numerator and denominator under the
same normalization as the main cache. Hard eviction is recovered when
$R=\varnothing$, where the second sums vanish and attention normalizes over
$M$ alone.

\subsection{Adaptive Residual Control}

\begin{figure}[t]
\centering
\includegraphics[width=\columnwidth]{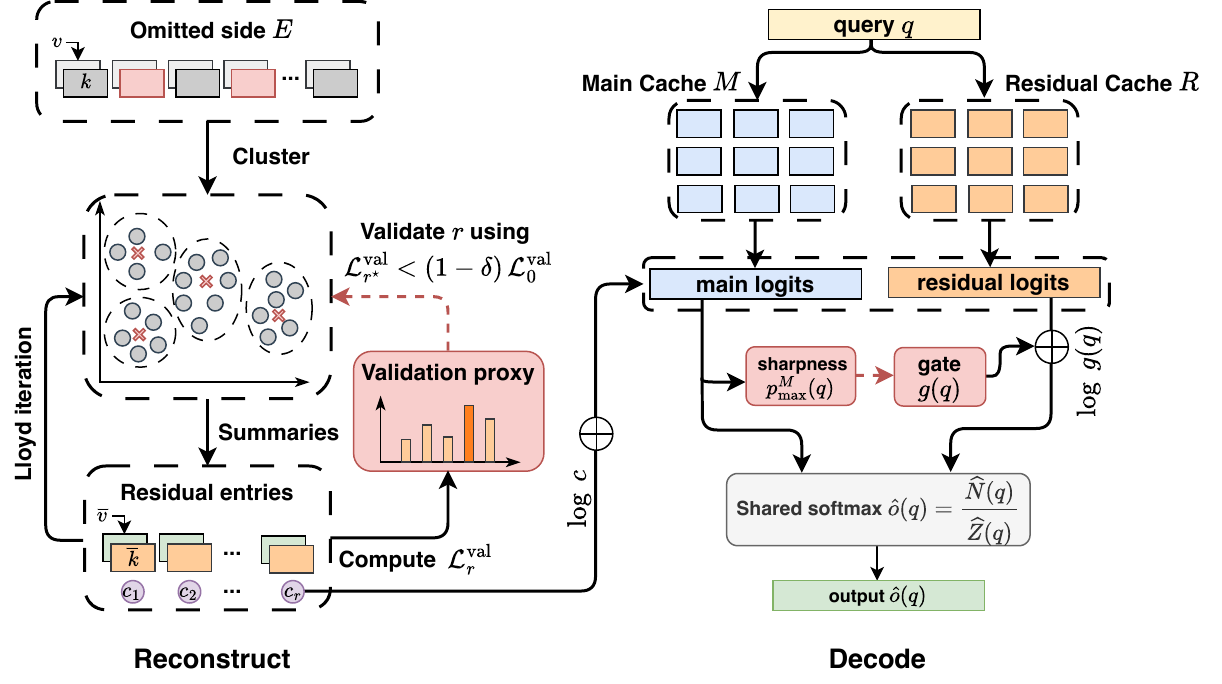}
\caption{Adaptive residual control. Left: the validation proxy selects the
residual allocation.
Right: the dynamic gate scales the residual logits according to the main-cache
sharpness before the shared softmax.}
\label{fig:residual_control}
\end{figure}

ResKV makes residual reconstruction adaptive rather than fixed. During cache
construction, a validation proxy selects the residual budget with an attention-output objective:
residual entries are allocated only when they improve held-out reconstruction of the
attention output. During decoding, the residual cache is used query-adaptively. Broad
attention patterns receive more residual mass to recover aggregate omitted
contributions, while sharp main-cache peaks are explicitly protected by down-weighting
the residual branch. These two controls align residual storage and residual usage with
the attention-output objective in Eq.~\ref{eq:residual_objective}.
Figure~\ref{fig:residual_control} illustrates the construction-time validation
proxy and the decode-time dynamic gate.

\subsubsection{Construction-Time Validation Proxy.}
The validation proxy selects the residual budget $r$ after prefill. We split the
observation window into a $\mathcal T_{\mathrm{fit}}$ part, used to compute token scores and select the
main cache, and a disjoint validation part $\mathcal T_{\mathrm{val}}$, used only to
judge the resulting cache. For each candidate $r$ in a small budget grid, ResKV sets
$m=b-r$, uses the fit scores to form $M$ under this budget as in
Eq.~\ref{eq:main_selection}, constructs residual entries from $E=S\setminus M$, and
evaluates the residual-augmented output
$\hat o^{(r)}(q_t)$ on validation queries. The validation loss is the reconstruction
error against the full-cache output:
\begin{equation}
\mathcal L^{\mathrm{val}}_r=\frac{1}{|\mathcal T_{\mathrm{val}}|}
\sum_{t\in\mathcal T_{\mathrm{val}}}
\big\lVert \hat o^{(r)}(q_t)-o_S(q_t)\big\rVert_2^2 .
\label{eq:valloss}
\end{equation}
The validation output uses the same residual-attention rule as decoding, including the
dynamic gate defined below. We take
$r^\star=\arg\min_{r>0}\mathcal L^{\mathrm{val}}_r$ and enable the residual only if it
beats the pure-main cache by a margin,
\begin{equation}
\mathcal L^{\mathrm{val}}_{r^\star}<(1-\delta)\,\mathcal L^{\mathrm{val}}_{0},
\label{eq:valproxy}
\end{equation}
where $\mathcal L^{\mathrm{val}}_0$ is the validation loss of the $r=0$ cache that
spends the whole budget on main tokens. Otherwise, the layer/head falls back to
$r=0$. This makes residual allocation layer- and KV-head-adaptive: residual entries
occupy budget only where they reduce held-out attention-output error.

\subsubsection{Decode-Time Dynamic Gate.}
When a query has diffuse attention over the main cache, the residual entries can
restore omitted aggregate mass. When the main cache already forms a sharp peak, the
averaged residual entries can add diffuse mass and dilute that peak. ResKV therefore
dynamically adjusts the residual contribution for each query.

\begin{table*}[t]
\centering
\caption{RULER average scores under different context lengths and retained KV budgets. Bold marks the better average within each baseline--ResKV pair. Complete task-level results are provided in Appendix.}
\scriptsize
\setlength{\tabcolsep}{2pt}
\renewcommand{\arraystretch}{0.90}
\resizebox{\textwidth}{!}{%
\begin{tabular}{@{}ll*{16}{c}@{}}
\specialrule{1.0pt}{0pt}{0pt}
\multirow{3}{*}{\textbf{Setting}} & \multirow{3}{*}{\textbf{Method}} &
\multicolumn{8}{c}{\textbf{LLaMA-3.1-8B}} &
\multicolumn{8}{c}{\textbf{Qwen-2.5-7B}} \\
\cmidrule(lr){3-10}\cmidrule(lr){11-18}
& & \multicolumn{4}{c}{\textbf{4K}} & \multicolumn{4}{c}{\textbf{32K}} & \multicolumn{4}{c}{\textbf{4K}} & \multicolumn{4}{c}{\textbf{32K}} \\
\cmidrule(lr){3-6}\cmidrule(lr){7-10}\cmidrule(lr){11-14}\cmidrule(lr){15-18}
& & \textbf{10\%} & \textbf{20\%} & \textbf{30\%} & \textbf{40\%} & \textbf{10\%} & \textbf{20\%} & \textbf{30\%} & \textbf{40\%} & \textbf{10\%} & \textbf{20\%} & \textbf{30\%} & \textbf{40\%} & \textbf{10\%} & \textbf{20\%} & \textbf{30\%} & \textbf{40\%} \\
\specialrule{0.9pt}{1pt}{1pt}
\rowcolor{gray!25}\multicolumn{2}{@{}l}{\textbf{Full Cache}} & \multicolumn{4}{c}{95.25} & \multicolumn{4}{c}{88.57} & \multicolumn{4}{c}{93.57} & \multicolumn{4}{c}{84.91} \\
\specialrule{0.9pt}{1pt}{1pt}
\multirow{4}{*}{\rotatebox[origin=c]{90}{\textbf{q-Aware}}} & AdaKV & 58.94 & 75.44 & 82.42 & 86.04 & 72.20 & 81.90 & 86.06 & 87.46 & 27.03 & 44.07 & 54.74 & 63.23 & 38.57 & 45.11 & 49.33 & 54.88 \\
& +ResKV & \textbf{70.51} & \textbf{83.34} & \textbf{86.37} & \textbf{88.34} & \textbf{77.84} & \textbf{86.14} & \textbf{87.83} & \textbf{88.51} & \textbf{28.31} & \textbf{44.84} & \textbf{57.37} & \textbf{63.92} & \textbf{39.85} & \textbf{46.04} & \textbf{49.69} & \textbf{55.01} \\
\cmidrule(lr){2-18}
 & SnapKV & 56.37 & 71.50 & 78.76 & 82.66 & 69.14 & 78.67 & \textbf{83.18} & 86.00 & 27.09 & 43.08 & 51.88 & 58.75 & 37.78 & 42.75 & 45.64 & 48.30 \\
& +ResKV & \textbf{60.74} & \textbf{75.59} & \textbf{80.10} & \textbf{83.31} & \textbf{70.31} & \textbf{78.92} & 83.14 & \textbf{86.50} & \textbf{28.50} & \textbf{45.70} & \textbf{55.64} & \textbf{60.39} & \textbf{38.79} & \textbf{43.74} & \textbf{46.25} & \textbf{48.62} \\
\specialrule{0.9pt}{1pt}{1pt}
\multirow{4}{*}{\rotatebox[origin=c]{90}{\textbf{q-Agnostic}}} & AdaKV & 21.52 & 39.11 & 50.55 & 60.74 & 26.92 & 38.23 & 52.57 & 64.27 & 18.38 & 26.38 & 33.41 & 39.43 & 25.91 & 29.45 & 33.89 & 36.81 \\
& +ResKV & \textbf{29.66} & \textbf{52.47} & \textbf{65.32} & \textbf{73.01} & \textbf{32.74} & \textbf{47.47} & \textbf{63.55} & \textbf{74.30} & \textbf{19.31} & \textbf{28.03} & \textbf{36.92} & \textbf{45.49} & \textbf{27.19} & \textbf{30.90} & \textbf{35.19} & \textbf{39.88} \\
\cmidrule(lr){2-18}
 & SnapKV & 19.38 & 34.23 & 44.47 & 53.45 & 24.45 & 31.07 & 38.59 & 47.66 & 16.81 & 25.42 & 30.56 & 34.99 & 25.09 & 28.77 & 31.27 & 32.99 \\
& +ResKV & \textbf{24.89} & \textbf{40.58} & \textbf{49.46} & \textbf{57.53} & \textbf{27.52} & \textbf{32.79} & \textbf{39.51} & \textbf{47.91} & \textbf{19.03} & \textbf{27.49} & \textbf{33.21} & \textbf{39.85} & \textbf{25.84} & \textbf{29.65} & \textbf{31.62} & \textbf{33.89} \\
\specialrule{1.0pt}{0pt}{0pt}
\end{tabular}%
}
\label{tab:ruler_paper}
\end{table*}

\begin{figure*}[t]
\centering
\includegraphics[width=\textwidth]{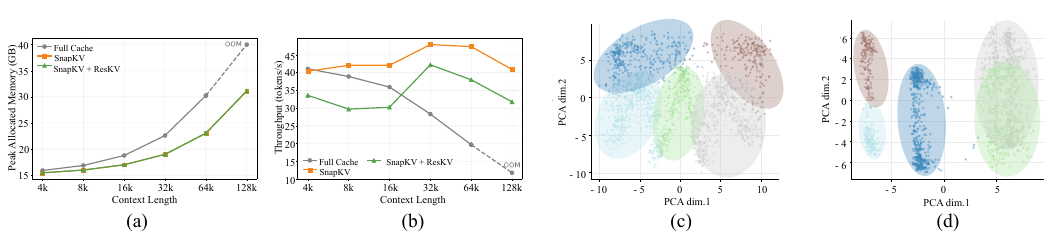}
\caption{Analysis of ResKV efficiency and residual clusterability. (a)--(b)
compare peak allocated memory and decode throughput across context lengths.
(c)--(d) show PCA projections of evicted candidate keys from
representative layers and KV heads after main-cache selection.}
\label{fig:analysis}
\end{figure*}

We measure main-cache sharpness by
\begin{equation}
p^{M}_{\max}(q)=\max_{p\in M}\frac{e^{a_p}}{\sum_{p'\in M} e^{a_{p'}}}
\end{equation}
the largest main-only attention weight. The dynamic gate is
\begin{align}
\tilde g(q)&=\operatorname{sigmoid}\!\left(\frac{\tau-p^{M}_{\max}(q)}{T_g}\right), \\
g(q)&=\max\!\left(g_{\min},\,\tilde g(q)\right),
\label{eq:gate}
\end{align}
This dynamic gate enters attention as the $\log g(q)$ shift in
\eqref{eq:residual_logit}. The temperature $T_g$ controls the smoothness of the
transition around the threshold $\tau$. A large $p^{M}_{\max}(q)$ indicates that the main cache
already provides a concentrated attention pattern, so the residual logits are
down-weighted. A smaller $p^{M}_{\max}(q)$ indicates broader main-cache attention, so
the residual entries retain more mass and can recover omitted aggregate contribution.

\section{Experiments}

\subsection{Experimental Setup}

\paragraph{Benchmarks.}
We evaluate ResKV on two long-context benchmarks: LongBench~\citep{bai2024longbench} and RULER~\citep{hsieh2024ruler}.
LongBench covers 16 real-world long-context understanding tasks, including
single-document QA, multi-document QA, summarization, few-shot learning,
synthetic retrieval, and code completion. RULER provides controlled
long-context tests for retrieval and aggregation; we report results at both 4K
and 32K context lengths over its 13 tasks.

\paragraph{Implementation Details.}
We evaluate ResKV on two instruction-tuned backbones,
LLaMA-3.1-8B-Instruct~\citep{grattafiori2024llama3} and
Qwen-2.5-7B-Instruct~\citep{yang2024qwen25}, under both query-aware and
query-agnostic cache construction at compression ratios
$\rho\in\{0.6,0.7,0.8,0.9\}$. ResKV uses the same total KV-slot budget as the
corresponding baseline, split between the main and residual caches. For all
methods, compression is performed once after prefill. Unless otherwise specified,
we use candidate residual budget ratios
$\{0,0.05,0.10,0.15,0.20\}$, 96 fitting queries, 32 validation queries, four Lloyd
iterations, $\delta=0.01$, and enable the dynamic gate at decode time.

Both backbones use grouped-query attention (GQA)~\citep{ainslie2023gqa}; ResKV
therefore constructs residual entries per KV head and shares them within the
corresponding query-head group. For FlashAttention-2~\citep{dao2024flashattention2},
we keep the main branch in the fused attention path and combine it with the residual
branch through log-sum-exp statistics, yielding the same shared softmax over main and
residual entries. All experiments are run on a single NVIDIA A100 GPU with 40GB
memory.

\paragraph{Baselines.}
We compare ResKV against the standard full KV cache and representative KV cache
compression baselines. H2O~\citep{zhang2023h2o} retains tokens with high accumulated
attention mass; SnapKV~\citep{li2024snapkv} estimates token importance from an
observation window; TOVA~\citep{oren2024tova} scores tokens using recent
attention; AdaKV~\citep{feng2025adakv} allocates cache budgets across
attention heads; and CaM~\citep{zhang2024cam} compresses the cache by merging
evicted states into retained entries. We report average tables for SnapKV
and AdaKV below, with complete task-level tables for other baselines and settings
provided in Appendix.

\subsection{Results on LongBench}
\begin{table}[t]
\centering
\caption{LongBench average scores under different retained KV budgets. Bold marks the better average within each baseline--ResKV pair. Complete task-level results are provided in Appendix.}
\scriptsize
\setlength{\tabcolsep}{2.2pt}
\renewcommand{\arraystretch}{0.96}
\resizebox{\columnwidth}{!}{%
\begin{tabular}{@{}ll*{8}{c}@{}}
\specialrule{1.0pt}{0pt}{0pt}
\multirow{2}{*}{\textbf{Setting}} & \multirow{2}{*}{\textbf{Method}} &
\multicolumn{4}{c}{\textbf{LLaMA-3.1-8B}} &
\multicolumn{4}{c}{\textbf{Qwen-2.5-7B}} \\
\cmidrule(lr){3-6}\cmidrule(lr){7-10}
 & & \textbf{10\%} & \textbf{20\%} & \textbf{30\%} & \textbf{40\%}
   & \textbf{10\%} & \textbf{20\%} & \textbf{30\%} & \textbf{40\%} \\
\specialrule{0.9pt}{2pt}{2pt}
\rowcolor{gray!25}
\multicolumn{2}{@{}l}{\textbf{Full Cache}} &
\multicolumn{4}{c}{48.93} & \multicolumn{4}{c}{48.14} \\
\specialrule{0.9pt}{2pt}{2pt}
\multirow{4}{*}{\rotatebox[origin=c]{90}{\textbf{q-Aware}}}
& AdaKV & 45.64 & 47.25 & 48.06 & 48.18 & 42.26 & 44.88 & 46.10 & 46.70 \\
& +ResKV & \textbf{47.39} & \textbf{48.45} & \textbf{48.71} & \textbf{48.81}
& \textbf{42.52} & \textbf{45.09} & \textbf{46.49} & \textbf{47.01} \\
\cmidrule(lr){2-10}
& SnapKV & 44.59 & 47.07 & 47.61 & 47.99 & 42.16 & 44.70 & 46.05 & 46.50 \\
& +ResKV & \textbf{45.66} & \textbf{47.52} & \textbf{48.06} & \textbf{48.44}
& \textbf{42.82} & \textbf{45.57} & \textbf{46.48} & \textbf{47.08} \\
\specialrule{0.9pt}{2pt}{2pt}
\multirow{4}{*}{\rotatebox[origin=c]{90}{\textbf{q-Agnostic}}}
& AdaKV & 36.26 & 42.10 & 44.78 & 45.95 & 33.28 & 38.86 & 41.97 & 43.84 \\
& +ResKV & \textbf{38.62} & \textbf{43.99} & \textbf{46.23} & \textbf{47.16}
& \textbf{35.63} & \textbf{41.45} & \textbf{43.18} & \textbf{44.95} \\
\cmidrule(lr){2-10}
& SnapKV & 34.79 & 40.63 & 43.67 & 45.17 & 33.22 & 39.02 & 42.08 & 43.98 \\
& +ResKV & \textbf{35.79} & \textbf{41.33} & \textbf{44.58} & \textbf{45.92}
& \textbf{35.24} & \textbf{40.51} & \textbf{42.64} & \textbf{44.60} \\
\specialrule{1.0pt}{0pt}{0pt}
\end{tabular}%
}
\label{tab:longbench_paper}
\end{table}

Table~\ref{tab:longbench_paper} reports the main LongBench average scores
across the evaluated settings. Complete task-level results and additional
baselines are provided in Appendix.

\subsubsection{Overall Performance.}
ResKV improves all 32 displayed LongBench configurations, with an average gain
of 1.02 points on LLaMA and Qwen. The gains are strongest
under tight budgets: at 10\% and 20\% retained KV, ResKV improves the
corresponding baseline by 1.43 and 1.17 points. Improvements are also larger in
the query-agnostic setting, which better matches practical serving because the
cache is compressed before the next user query is known.

\subsubsection{Task-Level Behavior.}
Task-level gains are most visible on code and retrieval-oriented tasks,
including RepoBench-P, LCC, Passage Retrieval, and TREC. These tasks often
depend on distributed definitions, repeated identifiers, or scattered evidence,
which can be missed by a small exact cache but recovered through residual
entries.

\subsection{Results on RULER}
Table~\ref{tab:ruler_paper} reports the main RULER average scores across
the evaluated settings. Complete task-level results and additional baselines
are provided in Appendix.

\subsubsection{Overall Performance.}
ResKV improves 63 of the 64 displayed RULER configurations, with an average
gain of 3.38 points across 4K and 32K contexts and both backbones.
The improvement is particularly clear in the query-agnostic setting, where the
cache is compressed before the future query is available: ResKV improves this
setting by 4.54 points on average, compared with 2.22 points under query-aware
construction.

\subsubsection{Effect Under Tight Cache Budgets.}
Under constrained cache budgets, ResKV improves the corresponding baseline by
3.47 and 3.66 points at 10\% and 20\% retained KV. Task-level gains concentrate
on controlled retrieval and aggregation behaviors, especially Variable Tracking,
FWE, and multi-key or single-key retrieval tasks, where useful evidence is
distributed across the context.

\subsection{Comparison with Merging Methods}
\begin{table}[t]
\centering
\caption{Task-level comparison under Qwen-2.5-7B-Instruct, 10\% retained KV, and query-agnostic cache construction.}
\scriptsize
\setlength{\tabcolsep}{3pt}
\renewcommand{\arraystretch}{0.95}
\resizebox{\columnwidth}{!}{%
\begin{tabular}{@{}lcccccc@{}}
\specialrule{1.0pt}{0pt}{0pt}
\textbf{Method} & \textbf{NrtvQA} & \textbf{HotpotQA} & \textbf{RB-P} & \textbf{FWE} & \textbf{S-1} & \textbf{QA-1} \\
\specialrule{0.9pt}{2pt}{2pt}
SnapKV & 19.43 & 36.48 & 56.82 & 57.07 & 19.20 & 20.80 \\
+CaM & 20.27 & 34.98 & 56.98 & 55.47 & 17.60 & 18.40 \\
+ResKV & \textbf{22.59} & \textbf{40.59} & \textbf{59.66} & \textbf{68.80} & \textbf{43.20} & \textbf{24.00} \\
\specialrule{1.0pt}{0pt}{0pt}
\end{tabular}%
}
\label{tab:snapkv_cam_reskv_tasks}
\end{table}

Table~\ref{tab:snapkv_cam_reskv_tasks} compares SnapKV, CaM, and ResKV under
10\% retained KV in the query-agnostic setting. CaM can recover useful omitted
information on some tasks, but its gains are less stable. ResKV gives stronger
gains by storing omitted information in a separate residual
cache while leaving the main cache unchanged.

\subsection{Analysis}

\paragraph{Peak Allocated Memory.}
Figure~\ref{fig:analysis}(a) shows that ResKV preserves the peak-memory
footprint of the compressed baseline. Its peak allocated memory essentially
overlaps with SnapKV across context lengths, indicating negligible additional memory
overhead, while the full cache grows rapidly and runs out of memory (OOM) at the
longest setting.

\paragraph{Decode Throughput.}
Figure~\ref{fig:analysis}(b) shows that ResKV incurs a moderate throughput cost
relative to SnapKV because each decode step evaluates the residual branch and
applies the dynamic gate. The throughput remains substantially higher and more
stable than the full cache at long context lengths; full-cache decode slows down
sharply and eventually OOMs, while ResKV maintains a stable decode rate through
128K context.

\paragraph{Clusterability of Evicted Keys.}
Figures~\ref{fig:analysis}(c)--(d) examine the keys evicted from the main cache
under SnapKV+ResKV. The projected keys show clear and repeated geometric
patterns rather than unstructured noise, showing that the omitted side still
contains organized key-space structure that residual entries can summarize.

\subsection{Ablation Study}

We conduct ablations with LLaMA-3.1-8B-Instruct using SnapKV as the base
method under 10\% retained KV and the query-agnostic construction
setting. We evaluate representative tasks from RULER-4K and LongBench, and
remove one component at a time from complete ResKV. Complete ResKV
performs best on all evaluated tasks.

\begin{table}[t]
\centering
\caption{Ablation results on representative tasks. Parentheses show drops from ResKV.}
\scriptsize
\setlength{\tabcolsep}{3pt}
\renewcommand{\arraystretch}{0.95}
\resizebox{\columnwidth}{!}{%
\begin{tabular}{@{}lcccc@{}}
\specialrule{1.0pt}{0pt}{0pt}
\textbf{Method} & \textbf{S1} & \textbf{VT} & \textbf{Gov.} & \textbf{Repo} \\
\specialrule{0.9pt}{2pt}{2pt}
\textbf{ResKV} & \textbf{67.20} & \textbf{45.92} & \textbf{25.33} & \textbf{50.86} \\
w/o validation & 63.20 {\scriptsize (-4.00)} & 38.88 {\scriptsize (-7.04)} & 25.09 {\scriptsize (-0.24)} & 50.66 {\scriptsize (-0.20)} \\
w/o dynamic gate & 62.40 {\scriptsize (-4.80)} & 41.44 {\scriptsize (-4.48)} & 25.13 {\scriptsize (-0.20)} & 50.41 {\scriptsize (-0.45)} \\
w/o shared softmax & 64.00 {\scriptsize (-3.20)} & 43.84 {\scriptsize (-2.08)} & 25.21 {\scriptsize (-0.12)} & 47.86 {\scriptsize (-3.00)} \\
\specialrule{1.0pt}{0pt}{0pt}
\end{tabular}%
}
\label{tab:ablation_main}
\end{table}

Removing the validation proxy reduces S1 and VT by 4.00 and 7.04 points, while
removing the dynamic gate reduces them by 4.80 and 4.48 points. Replacing the
shared softmax with separate main and residual normalizations reduces
RepoBench-P by 3.00 points. These drops show that all three components
contribute to the final performance.

\section{Conclusion}

We presented ResKV, a fixed-budget KV cache representation that preserves
selected tokens exactly in a main cache while reconstructing the aggregate
attention contribution of omitted tokens with compact residual entries. Its
shared-softmax residual decode places main-cache tokens and residual entries
under the same normalization, restoring residual numerator and denominator
mass while keeping omitted information separate from exact main-cache entries.
ResKV further controls residual usage
with a construction-time validation proxy and a decode-time dynamic gate.
Comprehensive evaluations on LongBench and RULER, spanning multiple backbones,
cache budgets, compression baselines, and both query-aware and query-agnostic
settings, show that ResKV improves all 32 displayed LongBench configurations
and 63 of the 64 displayed RULER configurations under the same retained KV
budget. The efficiency analysis further shows that these gains incur negligible
additional peak-memory overhead while maintaining stable decode throughput at
long context lengths.

\bibliography{refs}

\clearpage
\twocolumn
\appendix
\setcounter{secnumdepth}{3}
\renewcommand{\thesection}{\Alph{section}}
\renewcommand{\thesubsection}{\thesection.\arabic{subsection}}
\renewcommand{\thesubsubsection}{\thesubsection.\arabic{subsubsection}}
\makeatletter
\renewcommand{\@seccntformat}[1]{\csname the#1\endcsname.\quad}
\def\section{\@startsection {section}{1}{\z@}{-2.0ex plus -0.5ex minus -.2ex}{3pt plus 2pt minus 1pt}{\Large\bf\raggedright}}
\def\subsection{\@startsection{subsection}{2}{\z@}{-2.0ex plus -0.5ex minus -.2ex}{3pt plus 2pt minus 1pt}{\large\bf\raggedright}}
\def\subsubsection{\@startsection{subsubsection}{3}{\z@}{-1.5ex plus -0.5ex minus -.2ex}{3pt plus 2pt minus 1pt}{\normalsize\bf\raggedright}}
\makeatother

\section*{Appendix}

\section{Theoretical Analysis}

We provide additional derivations for the main-residual attention form used by
ResKV. The notation follows the preceding sections. All derivations are written
for one layer and one KV head unless otherwise stated.

\subsection{Main-Residual Decomposition}

This subsection shows why omitted cache statistics should be represented inside
the same softmax as retained tokens. The decomposition below separates the
full-cache output into main-cache and omitted-side numerator and denominator
terms.

For a query $q$, let $a_p(q)=\langle q,k_p\rangle/\sqrt d$. Following
Eq.~\eqref{eq:zn}, the unnormalized denominator and numerator over a token set
$\mathcal T$ are denoted by $Z_{\mathcal T}(q)$ and $N_{\mathcal T}(q)$. If the
main cache is $M$ and the omitted side is $E=S\setminus M$, full attention can
be decomposed as
\begin{equation}
Z_S(q)=Z_M(q)+Z_E(q),\qquad
N_S(q)=N_M(q)+N_E(q).
\end{equation}
The exact full-cache output is therefore
\begin{equation}
o_S(q)=\frac{N_M(q)+N_E(q)}{Z_M(q)+Z_E(q)} .
\label{eq:app_full_decomp}
\end{equation}
Hard eviction removes both omitted terms and returns $o_M(q)=N_M(q)/Z_M(q)$.
The resulting output difference is
\begin{equation}
o_S(q)-o_M(q)
=\frac{Z_M(q)N_E(q)-Z_E(q)N_M(q)}
       {Z_M(q)\left(Z_M(q)+Z_E(q)\right)} .
\label{eq:app_evict_error}
\end{equation}

\subsection{Residual Entry Approximation}

This subsection analyzes the approximation made by each residual entry. The
bound shows that a residual summary is accurate when grouped keys induce similar
query logits, and it becomes an exact KV token when the group size is one.

For a residual group $C_j\subset E$, the exact omitted denominator and numerator
are
\begin{equation}
Z_j(q)=\sum_{p\in C_j}e^{a_p(q)},\qquad
N_j(q)=\sum_{p\in C_j}e^{a_p(q)}v_p .
\end{equation}
ResKV stores the mean key $\bar k_j$, mean value $\bar v_j$, and count
$c_j=|C_j|$. Let
$\bar a_j(q)=\langle q,\bar k_j\rangle/\sqrt d$ and write the token logits in
the group as $a_p(q)=\bar a_j(q)+\epsilon_p(q)$. Then
\begin{equation}
Z_j(q)=e^{\bar a_j(q)}\sum_{p\in C_j}e^{\epsilon_p(q)} .
\end{equation}
The residual approximation replaces the group-level log-sum with
$c_j e^{\bar a_j(q)}$. When $|\epsilon_p(q)|\le \eta$ for all $p\in C_j$,
\begin{equation}
\left|Z_j(q)-c_j e^{\bar a_j(q)}\right|
\le c_j e^{\bar a_j(q)}\left(e^\eta-1\right).
\end{equation}
If the values in the same group satisfy $\|v_p\|_2\le V$, the numerator error is
bounded by
\begin{equation}
\left\|N_j(q)-c_j e^{\bar a_j(q)}\bar v_j\right\|_2
\le c_j e^{\bar a_j(q)}\left(e^\eta-1\right)V .
\end{equation}

\subsection{Validation Objective and Dynamic Gate}

This subsection describes the two control mechanisms used by ResKV. The
validation proxy selects layer- and KV-head-specific residual capacity under the
fixed retained KV budget, while the dynamic gate controls residual influence for
each decode query without changing main-cache logits or values.

For a candidate residual budget $r$, ResKV constructs a main cache and residual
cache, computes the shared-softmax output $\hat o_r(q)$ on validation queries,
and compares it against the full-cache output:
\begin{equation}
\mathcal{L}^{\mathrm{val}}_r
=\frac{1}{|\mathcal{T}_{\mathrm{val}}|}
  \sum_{q\in\mathcal{T}_{\mathrm{val}}}
  \|\hat o_r(q)-o_S(q)\|_2^2 .
\end{equation}
The residual budget is enabled only when this validation loss improves over the
pure-main cache by the relative margin $\delta$.

During decode, the dynamic gate modulates only residual logits. The main-cache
logits and values are left unchanged. With main-cache sharpness $p^M_{\max}(q)$,
the gate is
\begin{equation}
g(q)=\max\!\left(g_{\min},
\operatorname{sigmoid}\left(\frac{\tau-p^M_{\max}(q)}{T_g}\right)\right).
\end{equation}
A sharp main-cache distribution reduces residual mass, while a diffuse
main-cache distribution allows residual entries to contribute more strongly.

\section{Experimental Setup}

All experiments are run on a single NVIDIA A100 GPU with 40GB memory. The
implementation is based on PyTorch, Transformers, and FlashAttention-2. We
evaluate two instruction-tuned GQA backbones, LLaMA-3.1-8B-Instruct and
Qwen-2.5-7B-Instruct. All reported experiments are run once with a fixed random
seed of 42.

For GQA, ResKV constructs the main cache and residual entries at the KV-head
granularity. All query heads mapped to the same KV head share the selected main
indices and residual entries, while their logits and dynamic gates are computed
separately. For FlashAttention-2, the main-cache branch remains in the fused
attention path. We request the main-branch output and per-query log-sum-exp
normalizer, evaluate the residual branch over compact residual entries, and
merge the two branches through log-sum-exp statistics to produce one shared
softmax over main and residual entries.

\subsection{Benchmarks}

We evaluate on RULER and LongBench using their official benchmark instances.
The prompts, generation limits, answer formats, and metrics follow the
corresponding benchmark definitions without additional data preprocessing.
Across both benchmarks, we use greedy decoding and follow the benchmark metadata
for the number of generated tokens. For LongBench, we use each backbone's chat
template except for code-completion tasks, where the native completion format is
kept. For RULER, we use the same chat-template format for all tasks.

\paragraph{RULER.}
RULER evaluates controlled long-context behavior through synthetic retrieval and
aggregation tasks. We evaluate all 13 RULER tasks at 4K and 32K context lengths.
The official task templates, answer format, and string-match metric are used
without modification. The reported average is the unweighted mean over the 13
tasks.

\paragraph{LongBench.}
LongBench evaluates real-world long-context understanding over 16 tasks,
covering single-document QA, multi-document QA, summarization, few-shot
learning, synthetic retrieval, and code completion. The benchmark prompts,
answer prefixes, task-specific maximum generation lengths, and metrics are kept
unchanged. We report the official task-specific scores and use the unweighted
mean over the 16 tasks as the average.

\subsection{Baseline Setups}

\subsubsection{Prefill Compression}

This subsection fixes the common compression protocol used by all compared
methods. Compression is performed once after the prefill pass, and every paired
comparison uses the same total retained KV-slot budget under the same
compression ratio. It also defines the query-aware and query-agnostic settings,
where the latter mimics prefix-cache reuse before future user queries are known.

We evaluate both query-aware and query-agnostic construction. In the query-aware
setting, the downstream query is appended to the context before cache
construction. In the query-agnostic setting, the cache is constructed from the
reusable context alone, and the query is provided only after the compressed cache
has been formed.

We report compression ratios $\rho\in\{0.6,0.7,0.8,0.9\}$, corresponding to
retaining 40\%, 30\%, 20\%, and 10\% of the original KV slots. For each paired
comparison, ResKV uses the same base scoring or allocation rule as the
corresponding baseline and splits the same retained budget between main and
residual entries.

\subsubsection{Compared Methods}

This subsection lists the token scoring or budget allocation rule used by each
baseline.

\paragraph{H2O.}
H2O retains tokens with large accumulated attention mass. Following the
prefill-only protocol, scores are computed from observation queries after
prefill, and the most recent 64 tokens are protected.

\paragraph{TOVA.}
TOVA scores tokens using recent attention. We use the last-token observation
window and protect the most recent token. To keep the protocol consistent across
methods, TOVA selection is performed once after prefill rather than updated
during decode.

\paragraph{SnapKV.}
SnapKV estimates token importance from an observation window near the end of the
prefill sequence. We use an observation window of 64 tokens, apply average
pooling with kernel size 5, and protect the most recent 64 tokens.

\paragraph{AdaKV.}
AdaKV allocates the retained budget across KV heads. We use SnapKV scores as the
token-importance signal and set the safeguard coefficient to 0.2.

\subsection{ResKV Construction Settings}

Unless otherwise specified, ResKV uses the same construction settings across
benchmarks, backbones, baselines, cache budgets, and query settings. The
candidate residual budget ratios are $\{0,0.05,0.10,0.15,0.20\}$. For each layer
and KV head, ResKV uses 128 observation queries, with 96 fitting queries and 32
validation queries. The validation proxy enables residual entries only when the
held-out reconstruction loss improves over the pure-main cache by at least
$\delta=0.01$ in relative terms.

Residual construction uses four Lloyd iterations. The clustering initialization
is deterministic because, for each candidate residual budget, initial centers
are the evicted keys with the highest main-cache scores. Each residual entry
stores the mean key, mean value, and population count of one cluster. The
population count is included in the residual logit with exponent $\beta=1.0$.
The dynamic gate is enabled at decode time with main-cache sharpness threshold
$\tau=0.25$, sigmoid slope $\alpha=12.0$ (equivalently,
$T_g=1/\alpha$ in Eq.~\eqref{eq:gate}), and minimum residual gate
$g_{\min}=0.0$. Residual construction is enabled for all layers.

\section{Experimental Results}

The following tables provide the complete task-level results corresponding to
the average results reported earlier. Each table includes all evaluated retained
KV budgets and both construction settings. For every baseline--ResKV pair, bold
marks the better average score within the same backbone, benchmark, budget, and
construction setting.

\paragraph{RULER.}

Tables~\ref{tab:ruler4k_llama_full},~\ref{tab:ruler4k_qwen_full},
\ref{tab:ruler32k_llama_full}, and~\ref{tab:ruler32k_qwen_full} provide the
complete RULER results across the two evaluated context lengths and two
backbones. Each table reports all 13 RULER tasks and the unweighted average.

\begin{table*}[t]
\centering
\caption{RULER-4K results for LLaMA-3.1-8B-Instruct. Bold marks the better average within each paired setting.}
\label{tab:ruler4k_llama_full}
\setlength{\tabcolsep}{1pt}
\renewcommand{\arraystretch}{0.72}
\resizebox{0.88\textwidth}{!}{%
\begin{tabular}{@{}lll*{14}{c}@{}}
\specialrule{1.0pt}{0pt}{0pt}
Method & KV & Setting & CWE & FWE & MK1 & MK2 & MK3 & MQ & MV & S1 & S2 & S3 & QA-1 & QA-2 & VT & Avg. \\
\specialrule{0.8pt}{0pt}{1pt}
Full Cache &  &  & 99.60 & 85.07 & 100.00 & 100.00 & 100.00 & 100.00 & 100.00 & 100.00 & 100.00 & 100.00 & 88.00 & 65.60 & 100.00 & 95.25 \\
\cmidrule(lr){1-17}
\multirow{8}{*}{H2O} & 10\% & q-Aware & 65.20 & 42.67 & 29.60 & 8.80 & 0.00 & 34.60 & 7.00 & 44.80 & 44.80 & 1.60 & 79.20 & 59.20 & 17.92 & 33.49 \\
 & 10\% & q-Agnostic & 72.56 & 60.53 & 12.00 & 3.20 & 0.80 & 9.20 & 6.80 & 13.60 & 0.80 & 1.60 & 16.80 & 29.60 & 12.96 & 18.50 \\
 & 20\% & q-Aware & 86.32 & 48.80 & 85.60 & 28.00 & 16.80 & 74.60 & 17.20 & 75.20 & 79.20 & 1.60 & 88.00 & 63.20 & 40.16 & 54.21 \\
 & 20\% & q-Agnostic & 89.76 & 69.60 & 22.40 & 10.40 & 1.60 & 11.20 & 8.80 & 32.80 & 4.80 & 1.60 & 32.80 & 44.00 & 28.80 & 27.58 \\
 & 30\% & q-Aware & 94.96 & 56.80 & 99.20 & 57.60 & 34.40 & 91.00 & 41.00 & 83.20 & 94.40 & 1.60 & 88.80 & 64.00 & 65.76 & 67.13 \\
 & 30\% & q-Agnostic & 97.20 & 75.20 & 42.40 & 24.00 & 2.40 & 15.00 & 10.00 & 51.20 & 19.20 & 1.60 & 50.40 & 48.80 & 44.16 & 37.04 \\
\end{tabular}%
}
\end{table*}

\begin{table*}[t]
\centering
\caption{RULER-4K results for Qwen-2.5-7B-Instruct. Bold marks the better average within each paired setting.}
\label{tab:ruler4k_qwen_full}
\setlength{\tabcolsep}{1pt}
\renewcommand{\arraystretch}{0.72}
\resizebox{0.88\textwidth}{!}{%
\begin{tabular}{@{}lll*{14}{c}@{}}
\specialrule{1.0pt}{0pt}{0pt}
Method & KV & Setting & CWE & FWE & MK1 & MK2 & MK3 & MQ & MV & S1 & S2 & S3 & QA-1 & QA-2 & VT & Avg. \\
\specialrule{0.8pt}{0pt}{1pt}
Full Cache & -- & -- & 96.96 & 76.27 & 100.00 & 99.20 & 100.00 & 100.00 & 99.40 & 100.00 & 100.00 & 100.00 & 84.80 & 60.00 & 99.84 & 93.57 \\
\cmidrule(lr){1-17}
\multirow{8}{*}{H2O} & 10\% & q-Aware & 60.40 & 41.87 & 11.20 & 3.20 & 0.00 & 9.60 & 8.40 & 24.80 & 0.80 & 1.60 & 71.20 & 51.20 & 10.72 & 22.69 \\
 & 10\% & q-Agnostic & 52.56 & 60.53 & 11.20 & 1.60 & 0.00 & 8.00 & 8.40 & 16.00 & 0.80 & 1.60 & 22.40 & 24.80 & 5.12 & 16.39 \\
 & 20\% & q-Aware & 80.08 & 44.00 & 11.20 & 6.40 & 1.60 & 17.00 & 10.20 & 68.80 & 22.40 & 1.60 & 76.00 & 50.40 & 24.00 & 31.82 \\
 & 20\% & q-Agnostic & 75.20 & 65.87 & 11.20 & 2.40 & 0.00 & 9.40 & 8.60 & 28.00 & 0.80 & 1.60 & 40.00 & 35.20 & 8.00 & 22.02 \\
 & 30\% & q-Aware & 87.84 & 45.33 & 18.40 & 16.00 & 4.00 & 29.00 & 14.80 & 84.80 & 62.40 & 1.60 & 76.80 & 57.60 & 35.20 & 41.06 \\
 & 30\% & q-Agnostic & 86.48 & 65.33 & 11.20 & 4.80 & 0.80 & 9.40 & 8.80 & 48.00 & 2.40 & 1.60 & 47.20 & 40.80 & 12.16 & 26.07 \\
\end{tabular}%
}
\end{table*}

\begin{table*}[t]
\centering
\caption{RULER-32K results for LLaMA-3.1-8B-Instruct. Bold marks the better average within each paired setting.}
\label{tab:ruler32k_llama_full}
\setlength{\tabcolsep}{1pt}
\renewcommand{\arraystretch}{0.72}
\resizebox{0.88\textwidth}{!}{%
\begin{tabular}{@{}lll*{14}{c}@{}}
\specialrule{1.0pt}{0pt}{0pt}
Method & KV & Setting & CWE & FWE & MK1 & MK2 & MK3 & MQ & MV & S1 & S2 & S3 & QA-1 & QA-2 & VT & Avg. \\
\specialrule{0.8pt}{0pt}{1pt}
Full Cache &  &  & 31.44 & 88.27 & 100.00 & 98.40 & 99.20 & 99.80 & 99.40 & 100.00 & 100.00 & 100.00 & 84.00 & 52.00 & 98.88 & 88.57 \\
\cmidrule(lr){1-17}
\multirow{8}{*}{H2O} & 10\% & q-Aware & 22.72 & 69.60 & 92.80 & 20.00 & 15.20 & 91.80 & 51.60 & 98.40 & 96.80 & 4.00 & 81.60 & 50.40 & 84.64 & 59.97 \\
 & 10\% & q-Agnostic & 0.00 & 72.53 & 10.40 & 3.20 & 0.00 & 11.20 & 7.00 & 76.00 & 6.40 & 2.40 & 21.60 & 32.00 & 31.52 & 21.10 \\
 & 20\% & q-Aware & 31.36 & 79.73 & 99.20 & 43.20 & 55.20 & 99.00 & 91.60 & 100.00 & 96.80 & 20.00 & 81.60 & 51.20 & 94.24 & 72.55 \\
 & 20\% & q-Agnostic & 0.56 & 80.27 & 20.00 & 4.80 & 0.00 & 15.20 & 11.60 & 86.40 & 18.40 & 4.00 & 26.40 & 37.60 & 51.68 & 27.45 \\
 & 30\% & q-Aware & 31.68 & 83.20 & 99.20 & 58.40 & 76.80 & 100.00 & 97.00 & 99.20 & 98.40 & 47.20 & 82.40 & 51.20 & 96.16 & 78.53 \\
 & 30\% & q-Agnostic & 1.28 & 83.47 & 31.20 & 12.80 & 0.80 & 23.60 & 22.80 & 90.40 & 35.20 & 8.00 & 32.80 & 40.00 & 65.44 & 34.45 \\
\end{tabular}%
}
\end{table*}

\begin{table*}[t]
\centering
\caption{RULER-32K results for Qwen-2.5-7B-Instruct. Bold marks the better average within each paired setting.}
\label{tab:ruler32k_qwen_full}
\setlength{\tabcolsep}{1pt}
\renewcommand{\arraystretch}{0.72}
\resizebox{0.88\textwidth}{!}{%
\begin{tabular}{@{}lll*{14}{c}@{}}
\specialrule{1.0pt}{0pt}{0pt}
Method & KV & Setting & CWE & FWE & MK1 & MK2 & MK3 & MQ & MV & S1 & S2 & S3 & QA-1 & QA-2 & VT & Avg. \\
\specialrule{0.8pt}{0pt}{1pt}
Full Cache &  &  & 62.80 & 80.53 & 98.40 & 87.20 & 62.40 & 99.80 & 91.20 & 100.00 & 100.00 & 100.00 & 70.40 & 56.00 & 95.04 & 84.91 \\
\cmidrule(lr){1-17}
\multirow{8}{*}{H2O} & 10\% & q-Aware & 28.24 & 61.33 & 31.20 & 0.80 & 0.00 & 27.80 & 4.20 & 91.20 & 60.00 & 2.40 & 26.40 & 20.00 & 88.00 & \textbf{33.97} \\
 & 10\% & q-Agnostic & 31.68 & 61.60 & 8.80 & 2.40 & 0.00 & 9.60 & 6.40 & 68.00 & 0.80 & 2.40 & 16.80 & 28.80 & 24.16 & 20.11 \\
 & 20\% & q-Aware & 41.84 & 70.40 & 36.80 & 3.20 & 0.00 & 40.40 & 8.80 & 98.40 & 77.60 & 2.40 & 27.20 & 19.20 & 92.00 & 39.86 \\
 & 20\% & q-Agnostic & 40.24 & 66.40 & 8.80 & 4.80 & 0.00 & 10.20 & 6.20 & 83.20 & 0.80 & 2.40 & 20.00 & 32.80 & 52.00 & 25.22 \\
 & 30\% & q-Aware & 50.08 & 74.13 & 39.20 & 4.80 & 0.80 & 50.40 & 11.40 & 99.20 & 83.20 & 4.80 & 29.60 & 18.40 & 94.08 & 43.08 \\
 & 30\% & q-Agnostic & 44.08 & 71.73 & 8.80 & 5.60 & 0.80 & 10.20 & 6.40 & 91.20 & 0.80 & 2.40 & 24.00 & 39.20 & 63.68 & 28.38 \\
\end{tabular}%
}
\end{table*}

\paragraph{LongBench.}

Tables~\ref{tab:longbench_llama_full} and~\ref{tab:longbench_qwen_full}
provide the complete LongBench results for LLaMA-3.1-8B-Instruct and
Qwen-2.5-7B-Instruct. The tables report all 16 LongBench tasks and the
unweighted average.

\begin{table*}[t]
\centering
\caption{LongBench results for LLaMA-3.1-8B-Instruct. Bold marks the better average within each paired setting.}
\label{tab:longbench_llama_full}
\setlength{\tabcolsep}{2pt}
\renewcommand{\arraystretch}{0.92}
\resizebox{\textwidth}{!}{%
\begin{tabular}{@{}lll*{17}{c}@{}}
\specialrule{1.0pt}{0pt}{0pt}
Method & KV & Setting & \multicolumn{6}{c}{QA} & \multicolumn{3}{c}{Summ.} & \multicolumn{3}{c}{Few-shot} & \multicolumn{2}{c}{Synthetic} & \multicolumn{2}{c}{Code} & Avg. \\
\cmidrule(lr){4-9}\cmidrule(lr){10-12}\cmidrule(lr){13-15}\cmidrule(lr){16-17}\cmidrule(lr){18-19}
 & & & Narr. & Qasper & MFQA & Hotpot & 2Wiki & Musique & GovReport & QMSum & MultiNews & TREC & TriviaQA & SAMSum & PCount & PRe & LCC & RepoBench-P & Avg. \\
\specialrule{0.8pt}{0pt}{2pt}
Full Cache &  &  & 30.68 & 47.72 & 54.64 & 59.49 & 51.04 & 32.86 & 35.33 & 25.20 & 26.99 & 72.50 & 89.55 & 44.83 & 10.15 & 100.00 & 52.65 & 49.24 & 48.93 \\
\cmidrule(lr){1-20}
\multirow{8}{*}{H2O} & 10\% & q-Aware & 31.23 & 30.27 & 50.72 & 55.79 & 47.69 & 27.76 & 26.40 & 23.87 & 20.68 & 43.50 & 89.52 & 43.26 & 13.12 & 99.50 & 52.27 & 52.22 & 44.24 \\
 & 10\% & q-Agnostic & 25.82 & 15.29 & 26.14 & 41.79 & 24.05 & 18.02 & 25.38 & 19.85 & 20.90 & 44.00 & 90.93 & 43.38 & 9.50 & 50.00 & 52.27 & 47.91 & 34.70 \\
 & 20\% & q-Aware & 31.84 & 37.48 & 52.84 & 57.31 & 50.90 & 29.83 & 28.47 & 24.28 & 22.79 & 53.00 & 90.17 & 43.13 & 13.65 & 99.50 & 52.38 & 52.19 & 46.24 \\
 & 20\% & q-Agnostic & 28.66 & 23.39 & 35.57 & 49.63 & 36.15 & 21.25 & 28.05 & 21.20 & 22.48 & 48.50 & 90.52 & 44.21 & 10.00 & 84.00 & 52.38 & 48.83 & 40.30 \\
 & 30\% & q-Aware & 30.68 & 41.10 & 54.10 & 55.89 & 49.65 & 29.77 & 30.03 & 24.94 & 24.07 & 56.00 & 90.02 & 43.58 & 11.65 & 99.50 & 52.54 & 51.44 & 46.56 \\
\end{tabular}%
}
\end{table*}

\begin{table*}[t]
\centering
\caption{LongBench results for Qwen-2.5-7B-Instruct. Bold marks the better average within each paired setting.}
\label{tab:longbench_qwen_full}
\setlength{\tabcolsep}{2pt}
\renewcommand{\arraystretch}{0.92}
\resizebox{\textwidth}{!}{%
\begin{tabular}{@{}lll*{17}{c}@{}}
\specialrule{1.0pt}{0pt}{0pt}
Method & KV & Setting & \multicolumn{6}{c}{QA} & \multicolumn{3}{c}{Summ.} & \multicolumn{3}{c}{Few-shot} & \multicolumn{2}{c}{Synthetic} & \multicolumn{2}{c}{Code} & Avg. \\
\cmidrule(lr){4-9}\cmidrule(lr){10-12}\cmidrule(lr){13-15}\cmidrule(lr){16-17}\cmidrule(lr){18-19}
 & & & Narr. & Qasper & MFQA & Hotpot & 2Wiki & Musique & GovReport & QMSum & MultiNews & TREC & TriviaQA & SAMSum & PCount & PRe & LCC & RepoBench-P & Avg. \\
\specialrule{0.8pt}{0pt}{2pt}
Full Cache &  &  & 29.33 & 46.36 & 50.27 & 55.96 & 42.49 & 27.55 & 33.73 & 24.27 & 25.41 & 73.50 & 86.51 & 41.20 & 12.00 & 100.00 & 58.19 & 63.39 & 48.14 \\
\cmidrule(lr){1-20}
\multirow{8}{*}{H2O} & 10\% & q-Aware & 28.52 & 33.75 & 38.55 & 48.07 & 33.37 & 19.50 & 26.63 & 23.18 & 18.66 & 39.00 & 86.92 & 40.78 & 10.50 & 99.83 & 49.32 & 56.31 & 40.81 \\
 & 10\% & q-Agnostic & 20.28 & 13.38 & 24.14 & 31.35 & 23.28 & 14.87 & 26.30 & 18.17 & 19.36 & 42.50 & 86.89 & 39.26 & 8.00 & 54.92 & 49.54 & 57.79 & 33.13 \\
 & 20\% & q-Aware & 28.45 & 38.23 & 46.61 & 52.73 & 36.58 & 25.24 & 29.32 & 23.79 & 21.44 & 44.00 & 86.34 & 41.02 & 10.50 & 99.75 & 55.39 & 60.12 & 43.72 \\
 & 20\% & q-Agnostic & 23.83 & 20.25 & 29.15 & 40.58 & 27.65 & 17.40 & 28.96 & 19.52 & 21.64 & 46.00 & 85.59 & 39.98 & 6.00 & 91.08 & 55.41 & 59.74 & 38.30 \\
 & 30\% & q-Aware & 28.66 & 42.07 & 47.58 & 54.94 & 38.41 & 24.87 & 30.03 & 23.67 & 22.56 & 47.00 & 86.86 & 41.22 & 13.50 & 99.08 & 58.60 & 61.81 & 45.05 \\
\end{tabular}%
}
\end{table*}

\section{Implementation of ResKV}

We provide pseudo-code for ResKV in Algorithms~\ref{alg:reskv}
and~\ref{alg:decode}. Algorithm~\ref{alg:reskv} summarizes prefill-time cache
construction with residual-budget validation. Algorithm~\ref{alg:decode}
summarizes the decode-time shared-softmax computation with the dynamic residual
gate. Both algorithms are written for one layer and one KV head.

\begin{algorithm}[t]
\caption{ResKV construction for one (layer, KV head)}
\label{alg:reskv}
\textbf{Input}: $K,V$; budget $b$; residual grid $\mathcal{G}$; recent window $w$;
Lloyd steps $T$; margin $\delta$\\
\textbf{Output}: main cache $(K_M,V_M)$ and residual entries $\mathcal{S}$
\begin{algorithmic}[1]
\STATE form observation queries; split them into fit queries and
       $\mathcal{T}_{\mathrm{val}}$
\STATE compute token scores $s$ on the fit queries; set recent protection
       $\mathcal P$
\STATE build the pure-main cache ($r=0$) and compute $\mathcal L^{\mathrm{val}}_0$
\FOR{each $r\in\mathcal{G}$ with $r>0$}
  \STATE $m\leftarrow b-r$; select $M_r$ by \eqref{eq:main_selection}; set
         $E_r\leftarrow S\setminus M_r$
  \STATE build $r$ residual entries from $E_r$ by \eqref{eq:residual_partition}
         and \eqref{eq:summary}
  \STATE compute $\mathcal{L}^{\mathrm{val}}_r$ by \eqref{eq:valloss}
\ENDFOR
\STATE $r^\star\leftarrow\arg\min_{r>0}\mathcal{L}^{\mathrm{val}}_r$
\IF{$\mathcal{L}^{\mathrm{val}}_{r^\star}<(1-\delta)\mathcal{L}^{\mathrm{val}}_0$}
  \STATE choose $r^\star$
\ELSE
  \STATE choose $r=0$
\ENDIF
\STATE rebuild $(M,\mathcal S)$ with the chosen $r$; return $(K_M,V_M),\mathcal S$
\end{algorithmic}
\end{algorithm}

\begin{algorithm}[t]
\caption{ResKV decoding step for one (layer, KV head)}
\label{alg:decode}
\textbf{Input}: query $q$; main cache $(K_M,V_M)$; residual entries $\mathcal S$;
dynamic-gate parameters $\tau,T_g,g_{\min}$\\
\textbf{Output}: attention output $\hat o$
\begin{algorithmic}[1]
\STATE $a_p\leftarrow\langle q,k_p\rangle/\sqrt{d}$ for all $p\in M$
\STATE $p^{M}_{\max}\leftarrow\max_{p\in M} e^{a_p}/\sum_{p'\in M}e^{a_{p'}}$
\STATE $g\leftarrow\max\!\big(g_{\min},\,\operatorname{sigmoid}((\tau-p^{M}_{\max})/T_g)\big)$
\STATE $Z\leftarrow\sum_{p\in M} e^{a_p}$;\quad
       $N\leftarrow\sum_{p\in M} e^{a_p}v_p$
\FOR{each residual entry $(\bar k_j,\bar v_j,c_j)\in\mathcal{S}$}
  \STATE $\bar a_j\leftarrow\langle q,\bar k_j\rangle/\sqrt{d}+\log c_j+\log g$
  \STATE $Z\leftarrow Z+e^{\bar a_j}$;\quad
         $N\leftarrow N+e^{\bar a_j}\bar v_j$
\ENDFOR
\STATE \textbf{return} $\hat o\leftarrow N/Z$
\end{algorithmic}
\end{algorithm}

\section{Limitations and Future Work}

\paragraph{Broader model and serving coverage.}
ResKV is evaluated on two instruction-tuned GQA backbones and two long-context
benchmarks. Extending the evaluation to larger models, additional architectures,
and serving workloads with larger batch sizes would provide a broader view of
its practical behavior.

\paragraph{Residual updates during generation.}
The current construction is performed once after prefill, which matches
reusable prefix-cache serving. It does not refresh residual entries during long
generation. Future work can study lightweight residual updates for
generation-heavy workloads.

\paragraph{Residual summaries and kernels.}
ResKV uses mean key-value summaries selected by a validation proxy. Richer
residual summaries may further improve the approximation of omitted tokens.
The current implementation relies on standard PyTorch and FlashAttention-2
operations, while specialized kernels could further reduce the decode-time
overhead of the residual branch.

\end{document}